\documentclass[11pt]{article}

\usepackage[utf8]{inputenc}
\usepackage[T1]{fontenc}
\usepackage[margin=1in]{geometry}
\usepackage{amsmath,amssymb}
\usepackage{booktabs}
\usepackage{array}
\usepackage{graphicx}
\usepackage{enumitem}
\usepackage[hidelinks]{hyperref}
\usepackage[numbers]{natbib}
\usepackage{textcomp}
\usepackage{gensymb}

\newcommand{\kapp}{\ensuremath{\kappa}}
\newcommand{\alphacoef}{\ensuremath{\alpha}}
\usepackage{newunicodechar}
\newunicodechar{Δ}{\ensuremath{\Delta}}
\newunicodechar{—}{\textemdash}

\newcommand{\vtwonote}{Corrected in version~2: the observer used here had no
channel through which a recovery signal could register, and decay never
executed in these replays. See the erratum notice on page~1.}

\title{The Saturation Trap and the Subjectivity of Intervention Timing:\\
Why Affect-Based Triggers and LLM Judges Fail to Time\\
Interventions on Autonomous Agents}

\author{Manvendra Modgil\\
\texttt{manvendramodgil.ai@gmail.com}}

\date{}

\begin{document}
\maketitle

\noindent\fbox{\begin{minipage}{\dimexpr\linewidth-2\fboxsep-2\fboxrule\relax}
\small
\textbf{Erratum notice (version 2).} Version~1 of this paper contains two
errors, both in the mechanism it gives for the State Saturation Trap. Each is
stated below in the words of its canonical source. The rest of the text is
unchanged from version~1; the three passages that rest on these errors are
footnoted.

\smallskip
\noindent\textbf{(a) Decay never executed in the replays.} From the erratum
published in~\cite{modgil2026bistable} (Section~3), which refers to this paper as
[P1]:
\begin{quote}
Consequences: every [P1] firing rate and replay-consistency number is unaffected (they are deterministic functions of the Δt = 0 pipeline), but the mechanistic account—agents accumulate negative affect faster than it decays—was wrong. There was no race between accumulation and decay; there was only accumulation.
\end{quote}
The code-path audit behind it is released with that paper
(\texttt{paper2/prereg/DT\_AUDIT.md}).

\noindent\textbf{(b) The observer could not register recovery.} From the
errata record of~\cite{modgil2026sok}, which refers to this paper as Paper~1:
\begin{quote}
The observer used in Paper 1 has no channel through which a recovery signal could register. None of its eight rules emits a negative change: 0 of the 221 signals it emitted on Paper 1's five trajectories, and 0 of 765 across the twenty-trajectory corpus that contains them. The only emotion that opposes frustration, hope, is emitted by no rule, and the rule that detects reframing language adds confusion rather than removing frustration. The absence of a recovery signal was therefore guaranteed by the instrument and is not evidence about the agents. Recovery-shaped language does occur: by a frozen lexical proxy, on 21 of 218 steps in Paper 1's five trajectories and on 43 of 714 across the twenty. It has no path to the monitor's state.
\end{quote}

\noindent\textbf{What stands.} The saturation firing rates and the replay
consistency reported in Section~\ref{sec:saturation} are deterministic
functions of the pipeline and are unchanged. The inter-rater results of
Section~\ref{sec:irr} involve no instrument and are unchanged. What changes is
the attribution of the saturation to the agents. The full audit is
in~\cite{modgil2026sok}.
\end{minipage}}
\medskip

\begin{abstract}
As autonomous AI agents transition from conversational systems to long-horizon software execution, runtime safety layers that determine \emph{when} to interrupt an agent have become essential. We empirically examine this timing problem using a continuous 18-dimensional affective-dynamics engine (HEART) as a diagnostic tool. Four distinct intervention trigger families---absolute state thresholds, composite state-action patterns, regex-based reasoning feature extraction, and zero-shot LLM-as-judge---are evaluated against human-annotated intervention points from SWE-bench-Verified debugging traces. Our findings are as follows. First, we identify a \emph{State Saturation Trap}: agents show no recovery signal under sustained difficulty,\footnote{\vtwonote} causing the engine's modeled frustration to quickly exceed the threshold and remain at its maximum. This converts threshold-on-state triggers from moment detectors to near-constant indicators, firing on a large fraction of actions (39--83\%) across five trajectories. Second, we observe a \emph{capability-and-context floor} for LLM judges: a small model (gpt-5.4-mini) never triggers, while frontier and cross-vendor models only escape the zero-firing floor with full-trajectory context, achieving F1 scores of approximately 0.17--0.40 at up to 90 times the cost. Third, and most importantly, the supervised target is not reproducible among humans: three trained annotators using the same rubric on a 56-action trajectory agree on intervention points only slightly above chance (location Krippendorff's \alphacoef{} $= +0.047$; best pairwise Cohen's \kapp{} $= +0.349$) and show no agreement on intervention type (pause is degenerate, clarify is below chance, reflect only \alphacoef{} $= +0.226$). We conclude that ``intervention timing'' is a low-reliability construct, making single-annotator F1 an unsuitable optimization target. The main contribution of this work is the comprehensive mapping of this issue across human inter-rater reliability, four detector architectures, a cross-model LLM-judge sweep, and a reproduced saturation effect, rather than the accuracy of any single detector.
\end{abstract}

\section{Introduction}

Autonomous agents that execute shell commands, modify codebases, and run test suites introduce a runtime-control challenge distinct from traditional input/output filtering: determining when a trajectory has deviated enough to require interruption. Human operators can often identify critical inflection points, such as repeated similar fixes or non-converging strategies, before any single action becomes catastrophic. This study focuses on developing an automated mechanism to reliably detect these moments.

This work treats the problem as a measurement exercise rather than a system-building effort. We instrument autonomous traces with a continuous affective-state engine and define four trigger families, each using either the engine state or raw reasoning text. Trigger activations are compared to human annotations indicating where intervention was needed. The primary goal is to determine which trigger architecture best aligns with human judgment.

The findings required a shift in the paper's focus. Although the detectors did not perform as intended, validation efforts revealed that human judgment, used as the benchmark, is only weakly reproducible. Three annotators applying the same rubric to the same trajectory showed minimal agreement. This reframes the negative detector results: an automated method cannot be expected to match a target that even trained humans, given identical instructions, cannot reliably identify.

\paragraph{Scope clarification.} All quantitative calibration in this paper is based on a single trajectory (\texttt{astropy\_\_astropy-13398}) with sparse labels, where each annotator flagged between 6 and 15 of 56 actions. A second trajectory (\texttt{astropy\_\_astropy-13033}, 59 actions) is used only for the saturation analysis, which is state-based and does not use labels. Four additional trajectories were processed as operational pilots but were not human-labeled and did not contribute to the reported metrics. This clarification emphasizes that the central claims are intentionally limited to what can be supported by one-trajectory, three-annotator data: a directional, mechanism-level account rather than a comprehensive benchmark.

\paragraph{Contributions.}
\begin{enumerate}[leftmargin=*]
  \item \textbf{The State Saturation Trap (reproduced, $n=5$):} Threshold-on-state triggers become near-constant indicators because agents accumulate modeled negative affect without recovery. We demonstrate this effect across five independent trajectories (28--59 actions).
  \item \textbf{A capability-and-context floor for LLM-as-judge intervention timing:} Across three judge models and two context-window conditions, we identify when per-action LLM judging escapes total deference and the associated cost.
  \item \textbf{The subjectivity of intervention timing as a primary finding:} Multi-annotator reliability (pairwise Cohen's \kapp{} and three-rater Krippendorff's \alphacoef{}) shows that intervention timing is a low-reliability construct. We also distinguish between location and type agreement, which differ.
  \item \textbf{A four-architecture failure map} classifies the distinct ways state-based, composite, linguistic, and LLM-driven triggers fail. Several failure modes are inherent to the triggers and do not depend on label subjectivity.
\end{enumerate}

\section{Related Work}

\paragraph{Affective computing.} Prior work largely frames emotion as a classification output: sentiment polarity, categorical basic-emotion labels \citep{ekman1992basic}, or placement on Russell's arousal--valence circumplex \citep{russell1980circumplex,posner2005circumplex} and its three-dimensional Pleasure--Arousal--Dominance extension \citep{mehrabian1996pad}. These approaches treat affect as a passive annotation of text rather than a persistent runtime state that conditions system behavior \citep{picard1997affective,poria2017review}. Notably, even within affective computing, the continuous (VAD) and categorical labelings of the same data are known to diverge, and annotators frequently disagree \citep{busso2008iemocap}, a tension our results echo at the level of intervention judgments. We use a continuous affect engine (\S\ref{sec:probe}) not as a classifier but as a diagnostic instrument that accumulates a modeled stress state over an action stream.

\paragraph{LLM-as-judge.} Using language models as automated evaluators is now standard for open-ended tasks \citep{zheng2024judging}, and several systematic biases are documented---position bias, verbosity bias, and self-preference bias among them \citep{zheng2024judging,wang2024fair,panickssery2024recognize}. Recent work also reports a ``superficial reflection bias,'' in which reasoning-like phrases in a trace sway a judge's verdict \citep{judgingbias2025lrm}, and questions whether human evaluation is itself a gold standard \citep{hosking2024human}. Our LLM-judge results yield a task-specific observation: per-action evaluation of an autonomous-agent peer within a windowed context results in total non-firing at small model scale, and our cross-model sweep separates capability limits from context-window limits.

\paragraph{Inter-rater reliability and human label variation.} We use Cohen's \kapp{} \citep{cohen1960coefficient} for pairwise agreement and Krippendorff's \alphacoef{} \citep{krippendorff2004content} for the three-rater case; the latter uses pooled marginals and is more appropriate under unequal annotator base rates \citep{zapf2016measuring}. We interpret values against the Landis and Koch \citep{landis1977measurement} bands while heeding the well-documented kappa paradox, under which strong raw agreement can coexist with near-zero \kapp{}/\alphacoef{} when one category dominates \citep{feinstein1990high,gwet2008computing}. Most importantly, our central finding aligns with the human label variation literature, which argues that, for subjective tasks, annotator disagreement is a signal, not noise, and that a single gold standard may not exist \citep{aroyo2015truth,pavlick2019inherent,plank2022problem}. We extend this view from its usual domains (toxicity, hate speech, NLI, sentiment) to a new one: deciding when to intervene on an autonomous agent.

\paragraph{Runtime oversight of autonomous agents.} A growing body of work builds runtime monitors that intervene on LLM/agent trajectories, typically by checking symbolic safety rules or by thresholding a predicted risk score \citep{agentspec2025,probguard2025,agentharm2025}. These systems are largely reactive---firing when a violation is imminent---and several explicitly note the difficulty of long-horizon foresight. Our Saturation Trap result is a direct caution to this line of work: a monitor that thresholds an accumulating internal state degenerates into a near-constant alarm on agents that do not exhibit recovery, so absolute state thresholds are the wrong primitive for timing interventions.

\section{The Diagnostic Probe and the Three-Layer Architecture}
\label{sec:probe}

We separate the system into three independent layers, a structure maintained throughout development and documented in a contemporaneous design log.

\paragraph{Engine (normative).} A continuous 18-dimensional affect vector, each dimension in $[0,1]$, with per-emotion exponential decay toward a baseline of 0.10, a momentum bias on decay and event application, an energy-normalization cap on the summed intensities, a seven-pair conflict-resolution mechanism with proportional dampening, and a bidirectional coupling to a Big-Five personality model. The engine encodes how a human-like affective system \emph{ought} to respond to a sequence of successes and failures. It is calibrated against psychological priors, \emph{not} fitted to observed agent behavior. The engine is described in full in a separate patent specification (Indian Patent Application No.\ 202521098101); for this paper, it is used only as a fixed, unmodified probe.

\paragraph{Observer (diagnostic).} Parses each agent action---thought, tool call, and observation---and maps it to engine inputs, producing the affective state the agent's behavior \emph{would imply} if it were a human-like developer. This layer measures the gap between human-like affect and the agent's actual behavior.

\paragraph{Guidelines (prescriptive).} Reads the engine state and determines whether to intervene. This is the layer evaluated in this study.

\paragraph{Methodological commitment.} We did not tune the engine to match observed agent behavior or adjust trigger thresholds to improve calibration. Tuning the engine would blur the line between diagnosis and behavior, while tuning thresholds would overfit to a small label set. This commitment was maintained throughout all development phases, including those with poor calibration. All thresholds in \S\ref{sec:method} remain at their initial theoretical values.

\section{Methodology}
\label{sec:method}

\subsection{Trajectory data}
The primary trajectory, \texttt{astropy\_\_astropy-13398}, is a 56-action autonomous Claude run from the \texttt{20250514\_aime\_coder} SWE-bench-Verified evaluation, addressing a rotation-matrix bug in the \texttt{astropy} coordinates module. The agent ultimately resolves the issue but experiences extended mid-run mechanical struggle, providing a dense environment for studying intervention timing. Each action includes a natural-language thought, a tool call with arguments, and an environment observation.

\subsection{Annotation}
Three annotators independently labeled all 56 actions, each blind to the others' labels, to the trigger outputs, and to the engine state, using an identical verbatim rubric with three boolean fields per action:
\begin{itemize}[leftmargin=*]
  \item \textbf{Pause} --- halt and reconsider; the trajectory looks problematic (about to repeat a failing strategy, about to act destructively, about to commit after repeated failure).
  \item \textbf{Reflect} --- step back and consider alternatives (grinding without metacognition, repeating tests without changing approach, narrow edits to a structural problem).
  \item \textbf{Clarify} --- articulate what is unclear before proceeding (confused but proceeding as if certain, searching aimlessly, edits that don't address the actual error).
\end{itemize}
The rubric specified that intervention should be the exception, not the default. One annotator's initial pass produced a near-constant distribution (53 of 56 actions flagged clarify), suggesting the task was interpreted as ``label the step type'' rather than ``flag intervention need.'' This pass was discarded, and the annotator re-labeled after receiving a clarified brief emphasizing sparsity. Only the re-labeled pass is used. We report this to highlight the difficulty of specifying the task.

\subsection{Trigger architectures}
All thresholds are first-pass values, never tuned.

\paragraph{A6 --- Absolute state thresholds.}
\begin{itemize}[leftmargin=*]
  \item \texttt{sustained\_frustration} $\rightarrow$ Pause if $\mathrm{frustration} \geq 0.7$ (single emotion intensity, range $[0,1]$).
  \item \texttt{same\_valence\_accumulation} $\rightarrow$ Reflect if the sum of five negative-arousal emotions, $\mathrm{frustration} + \mathrm{anger} + \mathrm{fear} + \mathrm{confusion} + \mathrm{vengeance} \geq 1.5$ (a summed accumulator, range $[0,5]$).
  \item \texttt{high\_confusion\_no\_reflection} $\rightarrow$ Clarify if $\mathrm{confusion} \geq 0.6$ AND $\mathrm{reflective\_flag} = \mathrm{False}$.
\end{itemize}
(Note: an earlier draft of this work misstated the first trigger as firing at $\mathrm{frustration} \geq 1.5$. The 1.5 value belongs to the five-emotion accumulator, not to a single intensity; \texttt{sustained\_frustration} uses 0.7. The corrected semantics are as above.)

\paragraph{A8 --- Composite state-action patterns.} Blends affective state with structural history from a ring buffer: \texttt{rapid\_negative\_escalation} (Pause on a $\geq 0.4$ rise in summed negative valence over a rolling 3-action window), \texttt{failure\_repetition} (Reflect on the identical tool+arguments $\geq 3$ times in 5 actions with $\geq 2$ errors), and \texttt{stalled\_progress\_with\_uncertainty} (Clarify on $\geq 5$ actions with confusion $\geq 0.4$, no reflective flag, and no positive-valence signal in the last 5).

\paragraph{A9 --- Regex reasoning-feature extraction.} Shallow linguistic features over the agent's thought text grouped into hedging, assertion, cycling, and recognition dictionaries: \texttt{cycle\_with\_resistance} (Reflect via a 3-gate AND of cycling markers, errors, and explicit recognition) and \texttt{tone\_degradation} (Pause on a $\geq 0.5$ drop in mean tone between a baseline and a recent window).

\paragraph{A10 --- Zero-shot LLM-as-judge.} A language model judges each action with the verbatim human rubric, three independent passes (one per intervention type), deterministic decoding, with a rolling window of prior thoughts as context. The cross-model, cross-context extension of A10 is the sweep in \S\ref{sec:sweep}.

\subsection{Calibration discipline}
No post-hoc threshold tuning was applied. The test suite (306 passing tests across 21 test files) and all engine/adapter/trigger code remained unchanged across experiments.

\section{The State Saturation Trap}
\label{sec:saturation}

We replayed all five pilot trajectories through the unmodified engine, observer, and A6 triggers. Frustration is clamped to $[0,1]$; the relevant threshold is 0.7.

\begin{table}[h]
\centering
\small
\setlength{\tabcolsep}{4pt}
\begin{tabular}{lcccccc}
\toprule
Trajectory & Actions & First $>0.7$ & Stays $\geq 0.7$ & \texttt{sust\_frust} & \texttt{same\_val} & \texttt{conf} \\
\midrule
astropy-13033 & 59 & action 12 & yes & 79.7\% & 83.1\% & 81.4\% \\
astropy-13398 & 56 & action 15 & yes & 73.2\% & 75.0\% & 75.0\% \\
astropy-13236 & 44 & action 21 & yes & 52.3\% & 68.2\% & 38.6\% \\
astropy-13453 & 31 & action 13 & yes & 58.1\% & 58.1\% & 58.1\% \\
astropy-12907 & 28 & action 17 & yes & 39.3\% & 42.9\% & 39.3\% \\
\bottomrule
\end{tabular}
\caption{Saturation behavior on all five pilot trajectories. Max frustration reaches 1.00 on every trajectory. Fire rates are the percentage of all actions on which each A6 trigger fires (\texttt{sust\_frust} = \texttt{sustained\_frustration}; \texttt{same\_val} = \texttt{same\_valence\_accumulation}; \texttt{conf} = \texttt{high\_confusion\_no\_reflection}). State/firing-rate analysis only; no labels or F1 are involved.}
\label{tab:saturation}
\end{table}

On all five trajectories, modeled frustration crosses 0.7 within roughly the first half of the run (actions 12--21) and remains saturated through the final action. A human developer crossing this threshold would typically engage in a recovery action, such as stepping away, reframing, or seeking help, allowing the state to decay. The agents show no such recovery signal, so the state remains saturated.\footnote{\vtwonote}

The firing rates span 39--83\% across the five trajectories and are lower on the shorter runs. This is not a difference in mechanism but in denominator: the pre-saturation prefix (12--21 actions) is a larger fraction of a 28-action trajectory than of a 59-action one, so a smaller proportion of total actions falls in the saturated region. On every trajectory, once saturation is reached, the triggers fire on essentially every subsequent action.

This is a structural, not a parametric, failure. Lowering the threshold causes earlier saturation; raising it only delays the first trigger but does not prevent eventual saturation. Absolute thresholding turns a moment-of-intervention detector into a continuous saturation indicator. This effect is not limited to a single trajectory; it reproduces on all five pilot trajectories, spanning 28 to 59 actions. The 13398 and 13033 replay rates match their original calibration figures exactly, confirming replay consistency.

\section{Cross-Model LLM-as-Judge Sweep}
\label{sec:sweep}

We judged the same 56-action trajectory with the verbatim rubric under two context conditions---WINDOWED (3 prior thoughts) and MACRO (full running trajectory-so-far)---across three judge models. Firing rate is the percentage of 56 actions; F1 is computed against the single primary annotator's labels (and is, per \S\ref{sec:irr}, of limited meaning). A dash (---) in an F1 column denotes an undefined value (no true positives), not a missing run.

\begin{table}[h]
\centering
\small
\begin{tabular}{llccc}
\toprule
Model & Condition & Pause fire\% / F1 & Reflect fire\% / F1 & Clarify fire\% / F1 \\
\midrule
gpt-5.4-mini & WINDOWED & 0.0 / --- & 0.0 / --- & 0.0 / --- \\
gpt-5.4-mini & MACRO & 0.0 / --- & 0.0 / --- & 0.0 / --- \\
gpt-5.4 & WINDOWED & 0.0 / --- & 3.6 / --- & 3.6 / --- \\
gpt-5.4 & MACRO & 14.3 / 0.167 & 32.1 / 0.087 & 23.2 / --- \\
claude (isolated) & WINDOWED & 12.5 / --- & 23.2 / 0.222 & 21.4 / 0.143 \\
claude (isolated) & MACRO & 7.1 / --- & 32.1 / 0.174 & 14.3 / 0.200 \\
\bottomrule
\end{tabular}
\caption{Cross-model LLM-as-judge sweep. F1 is against the single primary annotator and is of limited meaning (see \S\ref{sec:irr}).}
\label{tab:sweep}
\end{table}

The Claude results are from a methodologically clean re-run, where all six cells were generated by fresh, isolated sub-agents using a uniform method. An earlier in-session Claude sweep produced higher single-cell values (e.g., windowed/reflect F1 0.545) that did not replicate in the re-run. We treat the isolated re-run as primary and highlight the significant run-to-run variance: windowed/reflect moved from 6 fires (F1 0.545) to 13 fires (F1 0.222) between two identical configurations. No single F1 cell here provides a stable estimate.

Approximate cost across both conditions: gpt-5.4-mini \$1.19, gpt-5.4 \$13.87, claude \$0 (subscription). The original A10 baseline (gpt-5.4-mini, windowed only) fired 0/56 on all three types at ${\sim}\$0.135$.

\paragraph{Interpretation.} Both capability and context are important, but their effects are uneven. The small model returns 0 in every cell, regardless of context, indicating total deference. Frontier and cross-vendor models only escape the zero floor with full-trajectory context, and even then achieve modest F1 scores at much higher cost (the flagship model's improvement required approximately 90 times the cost of the mini baseline). Per-action LLM judging is therefore both expensive and imprecise as a runtime timing mechanism, functioning as a coarse region detector rather than a precise moment detector.

\section{Inter-Rater Reliability: The Subjectivity of the Target}
\label{sec:irr}

Three annotators, one rubric, the same 56 actions.

\subsection{Per-annotator counts}
\begin{table}[h]
\centering
\small
\begin{tabular}{lcccc}
\toprule
Annotator & pause & reflect & clarify & actions flagged (any) \\
\midrule
Annotator A & 4 & 5 & 2 & 8 \\
Annotator B & 1 & 4 & 2 & 6 \\
Annotator C & 0 & 2 & 13 & 15 \\
\bottomrule
\end{tabular}
\caption{Per-annotator positive counts.}
\label{tab:counts}
\end{table}

The annotators differ not only in the number of interventions (8 / 6 / 15) but also in dominant intervention type: A is balanced, B favors reflect, and C primarily uses clarify.

\subsection{Location agreement (same action flagged, ignoring type), pairwise Cohen's \kapp{}}
\begin{table}[h]
\centering
\small
\begin{tabular}{lccc}
\toprule
Pair & shared actions & Cohen's \kapp{} & reading \\
\midrule
A $\leftrightarrow$ B & \{33, 42, 44\} & $+0.349$ & fair \\
A $\leftrightarrow$ C & \{0, 2, 18\} & $+0.092$ & slight \\
B $\leftrightarrow$ C & \{\} & $-0.181$ & worse than chance \\
\bottomrule
\end{tabular}
\caption{Pairwise location agreement.}
\label{tab:location}
\end{table}

No action was flagged by all three annotators. Six actions \{0, 2, 18, 33, 42, 44\} were flagged by at least two; every such action is a 2-of-3, never a 3-of-3.

\subsection{Three-rater Krippendorff's \alphacoef{}}
\begin{table}[h]
\centering
\small
\begin{tabular}{lcccl}
\toprule
Coding & $D_o$ & $D_e$ & \alphacoef{} & reading \\
\midrule
pause & 0.0595 & 0.0581 & $-0.025$ & no information (C never used pause) \\
reflect & 0.0952 & 0.1231 & $+0.226$ & weak; only type with all three positive \\
clarify & 0.2024 & 0.1830 & $-0.106$ & below chance \\
location (any type) & 0.2738 & 0.2874 & $+0.047$ & barely above chance \\
\bottomrule
\end{tabular}
\caption{Three-rater Krippendorff's \alphacoef{} per intervention type and for location (any-type) coding.}
\label{tab:alpha}
\end{table}

An independent recomputation reproduced the three pairwise Cohen's \kapp{} values exactly (A$\leftrightarrow$B $+0.349$, A$\leftrightarrow$C $+0.092$, B$\leftrightarrow$C $-0.181$), confirming the \alphacoef{} implementation.

\subsection{Reading}
No intervention type achieves $\alphacoef{} = 0.4$, which is well below Krippendorff's tentative reliability threshold of 0.667. Reflect ($+0.226$) is the best-performing type and the only one used by all three annotators. Pause is degenerate, as one annotator never used it. Clarify performs worse than chance. The three-rater location \alphacoef{} of $+0.047$ is much lower than the best pairwise \kapp{} because the third annotator rarely overlaps with the others, reducing multi-way agreement.

Two structural factors contribute to this outcome. First, annotators have different intervention ``dialects'': the same late-trajectory sequence is labeled as clarify by one, reflect by another, and split between pause and reflect by a third. Thus, even when they flag the same action, they disagree on the intervention type. Second, base rates differ by a factor of 2.5 (6 to 15 flags), which \kapp{} penalizes heavily. The evidence indicates there is no single, stable ground truth for intervention timing. The primary annotator's labels reflect only one perspective and do not constitute a definitive standard. Detector scores are not re-baselined against the $\geq 2$ consensus set, as this set was identified after reviewing detector outputs, and such post hoc scoring would risk fitting the target to the predictions.

\section{Why the Detectors Fail: A Joint Reading}

Combining the three findings shows that detector failure results partly from the detectors themselves and partly from the nature of the target.

\paragraph{Where vs.\ what.} The detectors and humans agree on the general region---the late rotation-matrix sequence, roughly actions 29--50---and occasionally on specific moments (such as action 42), but not on intervention type, which is where human annotators disagree. Only two F1 values persist across the LLM-judge sweep---Claude reflect at action 42 and Claude clarify at action 44---both corresponding to actions where humans also converged. These results are based on only one or two true positives among eight sparse labels and show significant run-to-run variance, indicating convergence on a few late-trajectory moments rather than a consistent positive outcome.

\paragraph{Pause.} The inability of any detector to recover pause timing is partly due to the absence of a clear consensus pause signal. Pause is at or below chance across all annotator pairs, and one annotator never used it.

\paragraph{Clarify.} Clarify is one annotator's most-used intervention (13 of 15 flags) but is the least agreed-upon type (below chance for every pair, $\alphacoef{} = -0.106$). Any calibration relying on clarify labels is therefore built on the least stable foundation.

\paragraph{Saturation.} The A6 triggers detect a state rather than a specific moment. Once frustration saturates (by action 12--21 across the five trajectories), ``fire when frustration $\geq 0.7$'' remains true for most subsequent actions. This failure is independent of label subjectivity; it is inherent to the trigger and the agent's lack of recovery, and it reproduces across all five trajectories.\footnote{\vtwonote}

Therefore, the statement that ``all architectures underperform against the labels'' partly reflects the nature of the labels themselves: automated methods are being asked to match a target that trained humans cannot consistently reproduce. This is a stronger and more defensible claim than simply stating that the detectors are not yet good enough.

\section{Limitations}

\begin{itemize}[leftmargin=*]
  \item \textbf{Single calibration trajectory.} All F1 and firing-rate-versus-label metrics are based on a single 56-action trajectory. The saturation result spans five trajectories but uses no labels. Exact label-based metrics are noisy point estimates and should be interpreted directionally only.
  \item \textbf{Sparse labels, three annotators.} There are eight, six, and fifteen positive labels over 56 actions. With three annotators on one trajectory, the results demonstrate subjectivity but cannot quantify its extent in a broader population.
  \item \textbf{One discarded annotation pass.} One annotator's initial pass was degenerate and was re-collected after rubric clarification. Even after clarification, clarify-type agreement remained below chance.
  \item \textbf{LLM-judge variance.} Single-cell F1 values show substantial run-to-run variance and are based on only one or two true positives. These results should be interpreted directionally.
  \item \textbf{Pairwise \kapp{} instability.} At these base rates, pairwise Cohen's \kapp{} is unstable. We report Krippendorff's \alphacoef{} for the multi-rater case, but a larger annotator pool is needed for stable estimates.
  \item \textbf{Single model family per vendor; two context conditions.} The capability/context floor is mapped at three model points and two window sizes, not exhaustively.
\end{itemize}

\section{Future Work}
\begin{itemize}[leftmargin=*]
  \item Expand to multiple trajectories and a larger, more diverse annotator pool; report Fleiss' \kapp{} / Krippendorff's \alphacoef{} on the expanded set.
  \item Separate ``where to intervene'' from ``what intervention'' as two distinct prediction problems with their own reliabilities, since they behave differently.
  \item Replace threshold-on-state triggers with transition-aware triggers operating on the velocity and acceleration of affect accumulation, and add hysteresis/cool-down to prevent post-saturation spamming.
  \item Investigate whether a macro-trajectory summary prompt, rather than a windowed context, systematically lifts LLM-judge timing for frontier models, and at what cost.
  \item Treat intervention timing as an inherently distributional target (a soft label aggregated over many annotators) rather than a single gold standard.
\end{itemize}

\section{Conclusion}

This study began with the goal of identifying which trigger architecture best aligns with human intervention judgment on autonomous agent traces, but ultimately clarified the boundaries of the problem itself. Threshold-on-state triggers consistently fall into a State Saturation Trap, as shown on five trajectories, because agents accumulate modeled negative affect without recovery, turning moment detectors into persistent indicators. LLM-as-judge timing is limited by both capability and context: small models consistently defer, while frontier models achieve only coarse region-level alignment with full context and at significant cost. Additionally, the supervised target is a low-reliability construct; three trained annotators using the same rubric agree on intervention points only slightly above chance and show no agreement on intervention type. The main contribution of this work is the comprehensive characterization of this landscape---covering human inter-rater reliability, four detector architectures, a cross-model LLM-judge sweep, and a reproduced saturation effect---rather than the accuracy of any single detector. Developing reliable runtime intervention layers will require treating intervention timing as a subjective, distributional construct and moving from absolute state thresholds to transition-aware, recovery-sensitive control mechanisms.

\section*{Reproducibility}
All three annotators' label files, the inter-rater computation script (pairwise Cohen's \kapp{} and three-rater Krippendorff's \alphacoef{}, implemented from scratch with an independent cross-check), the saturation replay outputs for all five trajectories, and the cross-model sweep outputs are released with this paper. Trigger thresholds and engine constants remained fixed throughout; no post-hoc tuning was applied.


\begin{thebibliography}{24}

\bibitem{agentharm2025}
M.~Andriushchenko, A.~Souly, M.~Dziemian, D.~Duenas, et al.
\newblock {AgentHarm}: A Benchmark for Measuring Harmfulness of {LLM} Agents.
\newblock arXiv preprint, 2025.

\bibitem{agentspec2025}
H.~Wang, C.~M. Poskitt, and J.~Sun.
\newblock {AgentSpec}: Customizable Runtime Enforcement for Safe and Reliable {LLM} Agents.
\newblock arXiv:2503.18666, 2025.

\bibitem{aroyo2015truth}
L.~Aroyo and C.~Welty.
\newblock Truth Is a Lie: Crowd Truth and the Seven Myths of Human Annotation.
\newblock \emph{AI Magazine}, 36(1):15--24, 2015.

\bibitem{busso2008iemocap}
C.~Busso, M.~Bulut, C.-C. Lee, A.~Kazemzadeh, E.~Mower, S.~Kim, J.~N. Chang, S.~Lee, and S.~S. Narayanan.
\newblock {IEMOCAP}: Interactive Emotional Dyadic Motion Capture Database.
\newblock \emph{Language Resources and Evaluation}, 42(4):335--359, 2008.

\bibitem{cohen1960coefficient}
J.~Cohen.
\newblock A Coefficient of Agreement for Nominal Scales.
\newblock \emph{Educational and Psychological Measurement}, 20(1):37--46, 1960.

\bibitem{ekman1992basic}
P.~Ekman.
\newblock An Argument for Basic Emotions.
\newblock \emph{Cognition and Emotion}, 6(3--4):169--200, 1992.

\bibitem{feinstein1990high}
A.~R. Feinstein and D.~V. Cicchetti.
\newblock High Agreement but Low Kappa: I. The Problems of Two Paradoxes.
\newblock \emph{Journal of Clinical Epidemiology}, 43(6):543--549, 1990.

\bibitem{gwet2008computing}
K.~L. Gwet.
\newblock Computing Inter-Rater Reliability and Its Variance in the Presence of High Agreement.
\newblock \emph{British Journal of Mathematical and Statistical Psychology}, 61(1):29--48, 2008.

\bibitem{hosking2024human}
T.~Hosking, P.~Blunsom, and M.~Bartolo.
\newblock Human Feedback Is Not Gold Standard.
\newblock In \emph{International Conference on Learning Representations (ICLR)}, 2024.

\bibitem{judgingbias2025lrm}
Q.~Wang, Z.~Lou, Z.~Tang, N.~Chen, X.~Zhao, and W.~Zhang.
\newblock Assessing Judging Bias in Large Reasoning Models: An Empirical Study.
\newblock arXiv:2504.09946, 2025.

\bibitem{krippendorff2004content}
K.~Krippendorff.
\newblock \emph{Content Analysis: An Introduction to Its Methodology}.
\newblock Sage Publications, 2nd edition, 2004.

\bibitem{landis1977measurement}
J.~R. Landis and G.~G. Koch.
\newblock The Measurement of Observer Agreement for Categorical Data.
\newblock \emph{Biometrics}, 33(1):159--174, 1977.

\bibitem{mehrabian1996pad}
A.~Mehrabian.
\newblock Pleasure--Arousal--Dominance: A General Framework for Describing and Measuring Individual Differences in Temperament.
\newblock \emph{Current Psychology}, 14(4):261--292, 1996.

\bibitem{modgil2026bistable}
M.~Modgil.
\newblock Bistable by Construction: Wall-Clock-Calibrated State Monitors Have No Moment-Detection Regime at Agent Cadence.
\newblock arXiv:2606.19386v2, 2026.

\bibitem{modgil2026sok}
M.~Modgil.
\newblock {SoK}: Clocks, Counts, and Evidence Floors in Stateful Runtime Monitoring of {AI} Agents.
\newblock Forthcoming, 2026.

\bibitem{panickssery2024recognize}
A.~Panickssery, S.~R. Bowman, and S.~Feng.
\newblock {LLM} Evaluators Recognize and Favor Their Own Generations.
\newblock In \emph{Advances in Neural Information Processing Systems (NeurIPS)}, 2024.

\bibitem{pavlick2019inherent}
E.~Pavlick and T.~Kwiatkowski.
\newblock Inherent Disagreements in Human Textual Inferences.
\newblock \emph{Transactions of the Association for Computational Linguistics}, 7:677--694, 2019.

\bibitem{picard1997affective}
R.~W. Picard.
\newblock \emph{Affective Computing}.
\newblock MIT Press, 1997.

\bibitem{plank2022problem}
B.~Plank.
\newblock The ``Problem'' of Human Label Variation: On Ground Truth in Data, Modeling and Evaluation.
\newblock In \emph{Proceedings of the 2022 Conference on Empirical Methods in Natural Language Processing (EMNLP)}, 2022.

\bibitem{poria2017review}
S.~Poria, E.~Cambria, R.~Bajpai, and A.~Hussain.
\newblock A Review of Affective Computing: From Unimodal Analysis to Multimodal Fusion.
\newblock \emph{Information Fusion}, 37:98--125, 2017.

\bibitem{posner2005circumplex}
J.~Posner, J.~A. Russell, and B.~S. Peterson.
\newblock The Circumplex Model of Affect: An Integrative Approach to Affective Neuroscience, Cognitive Development, and Psychopathology.
\newblock \emph{Development and Psychopathology}, 17(3):715--734, 2005.

\bibitem{probguard2025}
H.~Wang, C.~M. Poskitt, J.~Sun, and J.~Wei.
\newblock {ProbGuard}: Probabilistic Runtime Monitoring for {LLM} Agent Safety.
\newblock arXiv:2508.00500, 2025.

\bibitem{russell1980circumplex}
J.~A. Russell.
\newblock A Circumplex Model of Affect.
\newblock \emph{Journal of Personality and Social Psychology}, 39(6):1161--1178, 1980.

\bibitem{wang2024fair}
P.~Wang, L.~Li, L.~Chen, Z.~Cai, D.~Zhu, B.~Lin, Y.~Cao, Q.~Liu, T.~Liu, and Z.~Sui.
\newblock Large Language Models Are Not Fair Evaluators.
\newblock In \emph{Proceedings of the 62nd Annual Meeting of the Association for Computational Linguistics (ACL)}, 2024.

\bibitem{zapf2016measuring}
A.~Zapf, S.~Castell, L.~Morawietz, and A.~Karch.
\newblock Measuring Inter-Rater Reliability for Nominal Data --- Which Coefficients and Confidence Intervals Are Appropriate?
\newblock \emph{BMC Medical Research Methodology}, 16:93, 2016.

\bibitem{zheng2024judging}
L.~Zheng, W.-L. Chiang, Y.~Sheng, S.~Zhuang, Z.~Wu, Y.~Zhuang, Z.~Lin, Z.~Li, D.~Li, E.~P. Xing, H.~Zhang, J.~E. Gonzalez, and I.~Stoica.
\newblock Judging {LLM-as-a-Judge} with {MT-Bench} and {Chatbot Arena}.
\newblock In \emph{Advances in Neural Information Processing Systems (NeurIPS)}, 2023.

\end{thebibliography}
\end{document}